\documentclass[conference]{IEEEtran}
\IEEEoverridecommandlockouts
\usepackage{float} % this is another package for tables to introduce H sign.
\usepackage{cite}
\usepackage{amsmath,amssymb,amsfonts}
\usepackage{algorithmic}
\usepackage{graphicx}
\usepackage{textcomp}
\usepackage{xcolor}
\usepackage{comment}
\usepackage{booktabs} % for \specialrule{<thickness>}{<abovespace>}{<belowspace>}  command
\usepackage{multirow} % for multirow in table

\usepackage{tikz}
\usetikzlibrary{arrows,shapes,positioning,calc}
\tikzstyle{littlebox} = [rectangle, minimum width=0.7cm, minimum height=0.7cm, text centered, align=center, draw=black, fill=white]
\tikzstyle{nobox} = [rectangle, minimum width=0.7cm, minimum height=0.7cm, text centered, align=center, draw=white, fill=white]
\tikzstyle{arrow} = [very thick,->,>=stealth]
\tikzstyle{arrowLine} = [very thick]
\usepackage{array}
\usepackage{todonotes}
\usepackage{hhline}
\usepackage[normalem]{ulem}
\def\BibTeX{{\rm B\kern-.05em{\sc i\kern-.025em b}\kern-.08em
    T\kern-.1667em\lower.7ex\hbox{E}\kern-.125emX}}

\makeatletter
\newcommand{\linebreakand}{%
  \end{@IEEEauthorhalign}
  \hfill\mbox{}\par
  \mbox{}\hfill\begin{@IEEEauthorhalign}
}
\makeatother
    
\begin{document}

\title{Detection of Anomalies in Multivariate Time Series Using Ensemble Techniques\\

%\thanks{The work leading to these results has received funding from the European Union’s Horizon 2020 research and innovation programme under Grant Agreement No. 965231, project REBECCA (REsearch on BrEast Cancer induced chronic conditions supported by Causal Analysis of multi-source data).}
}

\author{\IEEEauthorblockN{Anastasios Iliopoulos}
\IEEEauthorblockA{Dept. Informatics \& Telematics \\
Harokopio University of Athens\\
Athens, Greece \\
itp20152@hua.gr}
%iliopoulosanastasis@gmail.com}
\and
\IEEEauthorblockN{John Violos}
\IEEEauthorblockA{Dept. Informatics \& Telematics \\
Harokopio University of Athens \\
Athens, Greece\\
violos@hua.gr}
\linebreakand
\IEEEauthorblockN{Christos Diou}
\IEEEauthorblockA{Dept. Informatics \& Telematics \\
Harokopio University of Athens \\
Athens, Greece\\
cdiou@hua.gr}
\and
\IEEEauthorblockN{Iraklis Varlamis}
\IEEEauthorblockA{Dept. Informatics \& Telematics \\
Harokopio University of Athens \\
Athens, Greece\\
varlamis@hua.gr}

}

\maketitle
%\begin{comment}...\end{comment}

\begin{abstract} %150-250 words
%\todo[inline]{301 words ==> Iraklis: 250 now} {250 words ==> anastasis: additions / 241 now}
%\todo{Anastasis: Please review because changed a little bit to include "nested-rotation" as proposed by @Christos Diou}
Anomaly Detection in multivariate time series is a major problem in many fields. 
Due to their nature, anomalies sparsely occur in real data, thus making the task of anomaly detection a challenging problem for classification algorithms to solve. 
% As the number of anomalies in real data is much smaller than the size of the normal data and there is a variety of anomaly types, the detection of anomalies becomes 
% a challenging problem for classification algorithms to solve. 
%Several 
Methods that are based on Deep Neural Networks such as LSTM, Autoencoders, Convolutional‐Autoencoders, etc., have shown positive results in such imbalanced data. However, the major challenge that algorithms face when applied to multivariate time series is that the anomaly can arise from a small subset of the feature set. 
%lies on the fact is that the anomaly can arise from a single feature or 
% a small subset of the 
%whole 
% feature set. 
To boost the performance of these base models, we propose a feature-bagging technique that considers only a subset of features at a time, and  we further apply a transformation that is based on nested rotation computed from Principal Component Analysis (PCA) to improve the effectiveness and generalization of the approach.
%and strengthen its robustness to noise. 
% PCA can introduce diversity and 
% generalization of the approach and strengthen its robustness to noise. 
%To further enhance the prediction performance, we propose an
%ensemble technique that combines the boosted basic models. 
To further enhance the prediction performance, we propose an ensemble technique that combines multiple base models toward the final decision.
% To further enhance the prediction performance, we propose an
% ensemble techniques that combine the basic algorithms. 
% The final result is an 
% ensemble model that combines the boosted basic models. 
% On the various basic models we applied the feature bagging and PCA. 
%More specifically, a Logistic Regressor is trained, in a semi-supervised setup, to learn how to combine the basic models.
In addition, a semi-supervised approach using a Logistic Regressor to combine the base models' outputs is proposed.
The proposed methodology is applied to the Skoltech Anomaly Benchmark (SKAB) dataset, which contains time series data related to the flow of water in a closed circuit, and the experimental results show that the proposed ensemble technique outperforms the basic algorithms.
%and their feature-bagging and PCA boosted alternatives. 
More specifically, the performance improvement in terms of anomaly detection accuracy reaches 2\% for the unsupervised and at least 10\% for the semi-supervised models.
%. On the other hand, the semi-supervised improvement is at least 10\%.
% of the ensemble is 10\% better than that of the basic algorithms. 
\end{abstract}


\section{Introduction}
%We should check if we describe all the important methods/models/approached we use i.e. the time series
A significant portion of actual data from systems, phenomena, or measurements is represented by time series. The ever-increasing volume of data makes it challenging for humans to monitor and analyze data, and for this, they rely on algorithms and machine learning models that automate such tasks and improve the monitoring process.
In the case of time series, machine learning models try to learn the patterns behind the evolution of a series and either predict future values or detect abnormal situations when they occur.
An anomaly is defined as an observation or sequence of observations that deviate significantly from the %general 
distribution of the data, and they usually constitute a very small portion of the total data \cite{braei2020anomaly}.
The anomalies are encountered whenever something goes wrong during the evolution of an operation, a phenomenon, or a process over time. It is crucial to identify these anomalous points and automatically determine if the corresponding sequence of values is an anomaly or not. Anomaly detection is linked to several real-life applications, such as the identification of fraudulent bank transactions, the early detection of machine malfunctioning, the detection of symptoms that indicate the existence of a disease or virus \cite{nassif_machine_2021}, and the detection of system intrusions \cite{mamalakis2019daemons}, just to mention a few. %A formal definition is as follows: An anomaly is an observation or sequence of observations that deviate significantly from the general distribution of the data and they constitute a very small portion of the total data \cite{braei2020anomaly}.
%The anomalies in a time series usually constitute a very small portion of the total data \cite{braei2020anomaly}. 

Anomaly detection can be approached mostly from two perspectives. It can be approached as a binary classification problem \cite{ullah_design_2022} that classifies an observation as an anomaly or not.
% where we have to answer the question of whether an observation as an anomaly or not. 
It can also be approached as an outlier detection problem where we seek unusual patterns or values that are far from the majority of observations \cite{gogoi2011survey}.
% it seems that an anomaly detection problem can be formulated as a standard classification problem. However, it is very far from it. 
This study follows the first approach and examines anomaly detection as a binary classification problem. However, apart from the fact that by definition anomalies are very sparse in a dataset, they can also vary significantly in type, which 
% anomalies themselves are different from each other based on their type. 
makes it very difficult for a standard classification algorithm to deal with. The three main anomaly types as described in \cite{chandolaBanerjeeKumar2009anomaly} are:
i) point anomalies, which refer to anomalies that have extreme values, ii) collective anomalies, which correspond to individual points that have common values with many other points, but which act very strangely as a group, and iii) context anomalies, which refer to points that in some environments or context are considered normal but in some other context are counted as anomalies.

% [We need sth here. We must highlight the research challenge we address and why we are different from others and what inspire us to propose our new model. We must say what already exists and what we introduce ]

%All these methods used in anomaly detection. 

While most anomaly detection models focus on a specific type of anomaly and neglect all other types, our proposed ensemble approach combines multiple models (weak models) that examine different types of anomaly and different features of the data.
% working on separate time series and  anomaly detection models, we propose the use of an
% our proposed ensemble approach combines multiple models (weak models) that examine different types of anomaly and different features of the data.
% work on different parts of the dataset. 
The ensemble methods we used combines the outputs of the weak models leveraging the diversity of their abilities to capture anomalies in multiple features concurrently. More specifically, in the context of this work, a multi-step  approach has been designed, implemented, and experimentally evaluated both in usupervised and semi-supervised setup. At first, multiple models, also named learners, are trained, each one using a different, randomly chosen, subset of features. Then a transformation is applied on each subset computed from principal component analysis (PCA) 
%is applied to every subset of features 
% reduce dimensionality while capturing 
to capture the variance of data and inject diversity. 

In the unsupervised setup the basic models are combined using the majority voting technique while in the semi-supervised alternative a Logistic Regressor is used to combine the predictions and provide a final answer.

The multi-step approach allows the detection of anomalies even when they are hidden in lower dimensions while at the same time, preserving the diversity of each learner. 
% Every learner has the ability to detect anomalies through a subset of features and by rotating its data it is different from the others.
As a result, the ensembles can perform well in higher dimensions and demonstrates an increased performance compared to the original methods.
% we manage to increase the performance while at the same time the model .

The major contributions of this paper are listed as follows:
\begin{itemize}
\item A method that applies Feature Bagging to the full set of features is combined with multiple basic anomaly detectors that specialize in different anomaly types and manages to uncover anomalies hidden in subsets of features.
\item A combination with a transformation based on Principal Component Analysis to the resulting feature sets further increases the diversity of detectors and better captures the different types of anomaly.
\item The two feature selection and transformation techniques, are combined with multiple detectors in an  ensemble model, which outperforms the baseline methods in terms of prediction performance.
\item the methodological approach proposed in this work is experimentally validated on a popular anomaly detection dataset.
\end{itemize}

The rest of this document is structured as follows: Section 2 briefly surveys the literature on anomaly detection techniques with an emphasis on ensemble methods and their application on time series. Section 3 briefly formulates the challenge that we address in this study. Section 4 formally defines the proposed approach and provides the details of the implemented model. Section 5 describes the experimental evaluation and provides an interpretation of the results. Finally, Section 6 concludes the paper and discusses the next steps of this work.

\section{Related Work}
%\textbf{[Anomaly detection categorizations based Surveys.]}\\
Anomaly detection techniques can be divided into six categories according to Cook \cite{8926446Cook}: statistical and probabilistic, pattern matching, distance-based, clustering-based, predictive, and finally ensemble. Additionally, Chandola et al \cite{chandolaBanerjeeKumar2009anomaly} divide the anomaly detection techniques into supervised, semi-supervised, and unsupervised based on the training mode employed in each case.
In a similar way,
% other works \cite{chandolaBanerjeeKumar2009anomaly, chandola2009anomaly} categorize 
the methods of the literature are categorized into three main groups: statistical methods, machine learning methods, and deep learning methods \cite{chandolaBanerjeeKumar2009anomaly}. In these categorizations, statistical methods assume that anomalies are generated from a statistical model. In contrast, in machine learning methods, the anomaly generation mechanism is considered a black box and anomalies are detected based on the data. Finally, the third category comprises techniques that are based exclusively on neural networks. They are often considered part of machine learning so there is often no separation between the second and third categories.

\begin{figure}[htbp]
\centerline
{\includegraphics[width=\columnwidth]{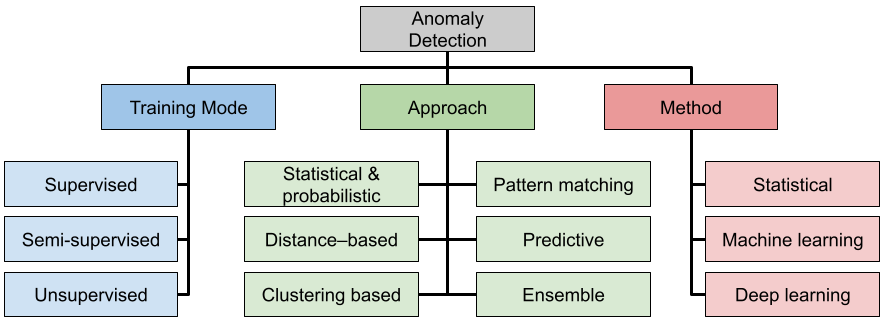}}
\caption{Alternative groupings of anomaly detection techniques.}
\label{fig:methods}
\end{figure}

Several methods in the literature employ statistical \cite{Singh2003Novelty} and machine learning techniques \cite{markou2003novelty} to detect anomalies in multivariate time series. The works that build on statistics rely on assumptions about the distribution of data, but their advantage is the cheap computation and the straightforwardness in explaining them. 
On the other hand, machine learning models can generalize better, but they have more computationally expensive training and retraining in case new datasets arrive. The respective works list several machine learning approaches based on Multi-layer perceptrons, Support vector machines, and Auto-associator approaches such as the PCA. 

A grouping of the alternative techniques is depicted in Figure~\ref{fig:methods}.
%\textbf{??????[Position your work into this research topic.]??????}
The methodology we propose 
% and will be described in the following sections
is based on machine learning techniques and combines elements both from unsupervised and semi-supervised anomaly detection. It can be considered an ensemble deep learning model as it uses a variety of neural networks as a base and builds upon them.
The ensemble model makes use of the basic neural networks to discover sequential and spatial relationships in time series.
Concerning the nature and distribution of data, we do not make any special assumptions.
% , and 
% % about the distribution or the general nature of data. So we 
% consider the data generation mechanism as a black box. 

%\textbf{[Mention Baseline (optional) and State of the art in Anomaly detection methods. You will use some of these in your experimental evaluation]}
In the related literature, we encounter methods that are close to our proposed model, so in the following, we discuss their similarities and differences and use them in the experimental comparison.
Chauhan and Vig \cite{chauhan2015anomaly} use the Long Short Term Memory (LSTM) network \cite{hochreiter1997long} on an electrocardiography signals dataset that contains 1-dimension time series to detect anomalies. Malhotra et al. \cite{mukhopadhyay2020fault} use CNNs on a balanced dataset of healthy and broken tooth gears, to detect anomalies. Autoencoders have been first introduced in \cite{rumelhart1985learning} and  have also been extended to detect anomalies in \cite{malhotra2016lstm}. Malhotra in \cite{malhotra2016lstm} uses a network similar to the LSTM Autoencoder on three 1-dimension datasets and one proprietary multiple-dimension dataset. 
% The same can be applied with a CNN as a decoder and encoder. 
Park in \cite{park2018multimodal} uses an LSTM-Variational Autoencoder (VAE) model for anomaly detection problems on multidimensional time series. In \cite{skabdataset} authors have built a collection of multivariate time series, which is used to evaluate anomaly detection algorithms and provide a variety of algorithms as a baseline. Finally, in \cite{elhalwagy2022hybridization} a method is proposed that results in a hybrid network using LSTM and Capsule networks, which are evaluated on the SKAB dataset.

%\textbf{[Say the limitations of the other methods]}
Although many of the above-mentioned methods are available to detect anomalies in \cite{skabdataset} we see that the performance they achieve is not satisfactory and there is significant room for improvement. In this direction, authors in \cite{lazarevicKumar2005Featurebagging} suggest the selection of subsets of the feature set to take advantage of the fact that anomalies in high dimensional data are hardly detected in all the dimensions 
% may be found under a subset of the feature set, and 
% only a subset of features could clearly reflect the anomalous behavior and 
and can better be detected using only a subset of features. In the same direction, the authors in \cite{Rotation2006Forest} employ  ensembles 
%for anomaly detection 
and boost their performance by increasing the diversity of the learners they employ.
% plays an important role in performance of the ensemble techniques. 
For this purpose, they propose a technique called Rotation Forest to increase diversity and accuracy together.

%\textbf{[Here we should say that as the ML methods are evolving, ensemble methods are emerged in order to combine the advantages and the different aspects of data.]}
As machine learning methods are evolving, ensemble methods emerge, aiming to combine the advantages of the individual models and methods to capture the different aspects of data. However, ensemble methods are not yet widely used in anomaly detection \cite{aggarwal2013outlier, zimek2014ensembles}. Furthermore, the authors in \cite{chiang2017study}  concluded that anomaly detection using an ensemble of methods and an unsupervised approach has limited value, and suggested the use of ensemble methods in semi-supervised or supervised approaches. 
Therefore, to address the limitations of the above-mentioned approaches and increase the performance of the anomaly detection task, we first recommend 
% Addition 
the use of both Feature Bagging \cite{lazarevicKumar2005Featurebagging} and a transformation based on Rotations \cite{Rotation2006Forest} which we call Nested Rotations, simultaneously in an unsupervised setup. To further extend the prediction performance we also evaluate the use of an ensemble technique that follows a semi-supervised approach using the aforementioned techniques. 
% \textcolor{blue}{With the 
The Feature Bagging technique allows us to shrink the feature space and predict anomalies based only on a subset of features. With the transformation based on Nested Rotations, we divide the problem sub-space into partitions and rotate them to increase the data diversity.
%With the Rotation technique, we attempt to rotate the problem sub-space and increase the data diversity. 
For this purpose, we employ the PCA method which gives us orthogonal eigenvectors to use as our new axes.

\section{Problem formulation}
Let $X=(X_t : t \in T)$ be a multivariate time series, where $T$ is the index set and $X_t \in \mathbb{R}^d, \forall t \in T$. For all $t \in T$ we assign a score which is called anomaly score. Let $\text{IQR}$ be the $\text{IQR}$ of all anomaly scores the we get during the training phase then, based on a threshold $\delta = 1.5 * \text{IQR}$ it is decided if $X_t$ is an anomaly when the score is greater than $ \delta$ or not an anomaly otherwise. So we denote the anomaly score of $X_t$ as $\text{ASc}(X_t)$ and transform this problem into a binary problem by defining $\text{ASc}_\text{binary}(X_t)=1$ if $\text{ASc}(X_t) > \delta$ and $\text{ASc}_\text{binary}(X_t)=0$ otherwise. Every detector has a different scale for anomaly score for a given $X_t$ but all detectors have binary outcomes ($\text{ASc}_\text{binary}(X_t)$) when we define a threshold. Our approach defines multiple detectors so let $\text{ASc}^i_\text{binary}(X_t)$ be the binary outcome for the $i$-th detector for thr $X_t$ instance. Then we could aggregate the anomaly scores of detectors into one $ASc(X_t)=\text{agg}_{\forall i}(ASc^i(X_t))$ and then transform it into binary. The alternative approach, and the one that we use in this study, is to aggregate the binary outcomes into one binary outcome $ASc_\text{binary}(X_t)=\text{agg}_{\forall i}(ASc^i_\text{binary}(X_t))$.

\section{Proposed approach}

%is to take 5 architectures (Autoencoder, Convolutional Autoencoder, LSTM, LSTM Autoencoder and LSTM VAE). In this study we propose an ensemble model to tackle the problem of anomaly detection in time series data.  two individual methods which are combined into one general model. %The main idea of the general model is that by applying the two individual techniques, 

\begin{figure}[htbp]
\centerline{\includegraphics[width=0.5\textwidth]{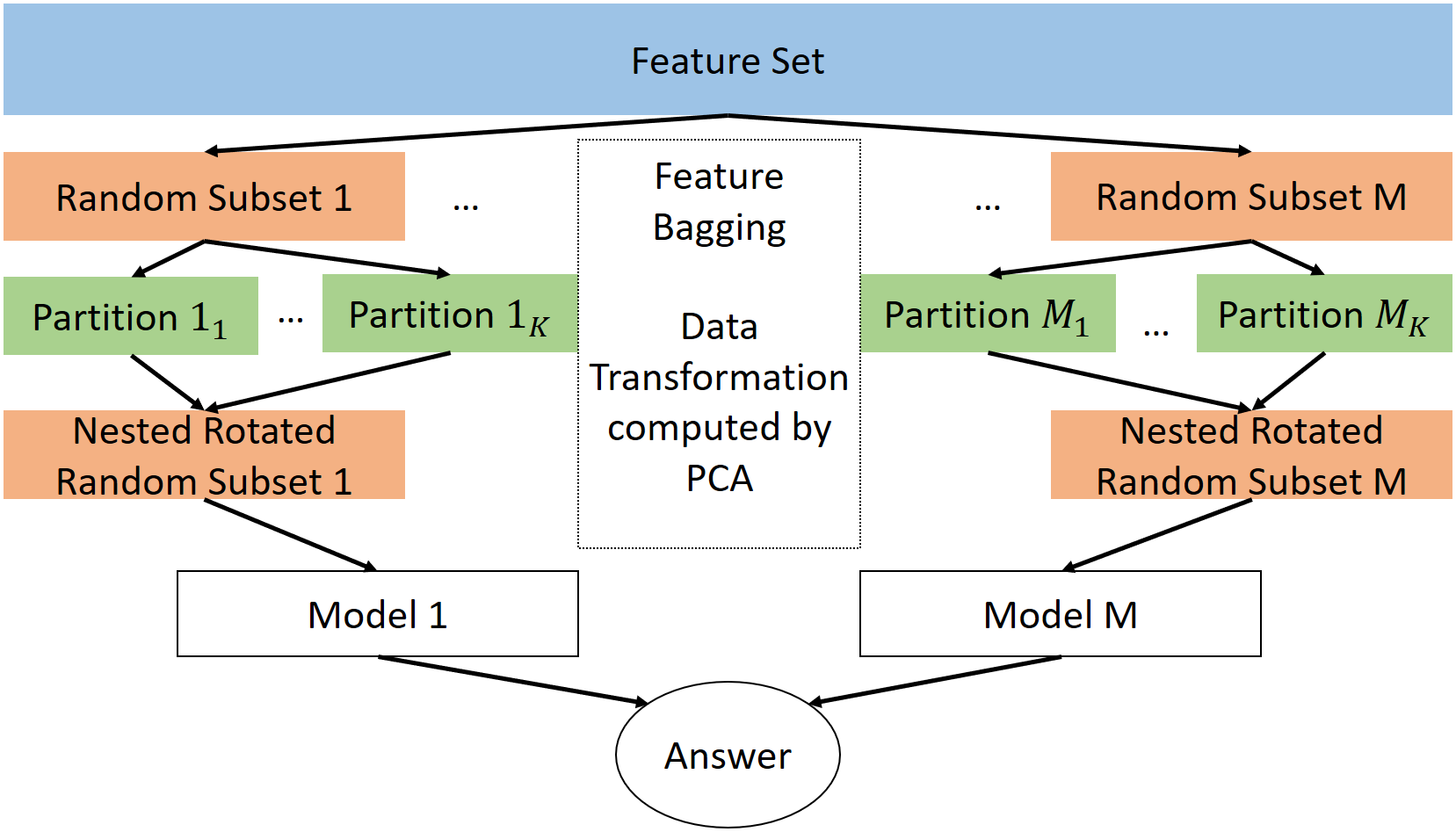}}
\caption{Feature Bagging \& Nested Rotations}
\label{fig:fbr}
\end{figure}

\begin{figure}[htbp]
\centerline{\includegraphics[width=0.5\textwidth]{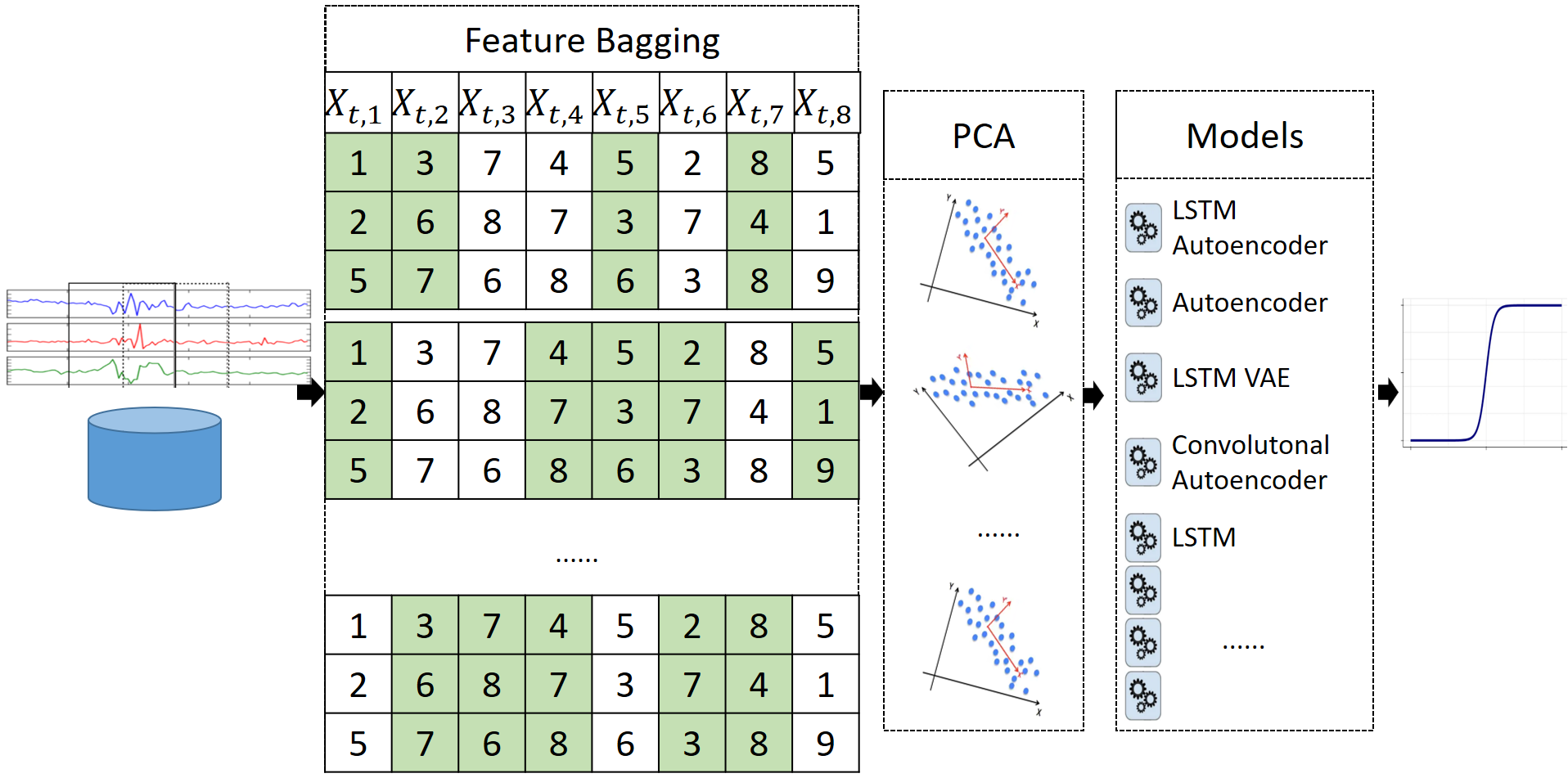}}
\caption{Stacking/Fusion Feature Bagging with Nested Rotations models with Logistic Regression}
\label{fig:fig_stacking_fbr_log_reg}
\end{figure}

As explained above, we approach the anomaly detection problem in multivariate time series as a binary classification problem and rely on an ensemble model and feature engineering to improve the classification performance. We assume that the distribution of data is unknown and thus we depend on a representation learning approach to capture the patterns hidden in the multivariate time-series data. For this purpose, we apply two feature aggregation and a transformation techniques, namely Feature Bagging and transformation based on Nested Rotation (computed by applying PCA), on five different base architectures (i.e. Autoencoder, Convolutional Autoencoder, LSTM, LSTM Autoencoder and LSTM Variational Autoencoder) in the fully unsupervised approach. In the semi-supervised approach, the predictions of the resulting models are combined with the help of a Logistic Regressor, which is trained to create an efficient ensemble. As depicted in Figures \ref{fig:fbr} and \ref{fig:fig_stacking_fbr_log_reg}, it is important to decide on the number of models to use in the ensemble both in unsupervised learning and semi-supervised learning, and this is affected by the number of base models, the different feature subsets and the rotations applied to them. In the first step of the semi-supervised approach, the base architectures that will be used to develop the prediction models must be carefully selected to capture the intrinsic characteristics of the various time series. In both setups (unsupervised and semi-supervised) it is important to 
% firstly we need to choose how many models of each architecture we need to create. Next we 
apply the Feature Bagging technique on the dataset and create several subsets, each one comprising a subset of the original feature, that will be used to train the respective models.
The next step is the Nested Rotations of each subspace (defined by the respective subset of features), which is a transformation performed using PCA on partitions of each subspace.
% set in order to create several data sets, one for each model. The second technique is the Rotation which it takes every set of the defined sets by the previous technique and divide it into further subsets. On these subsets we apply the PCA algorithm to rotate the space and connect them back to form the initial set but now they are rotated. 
The final step in the semi-supervised version is the training of the models with the transformed subsets and their integration in an ensemble predictor, based on 
% that created by the previous steps with these data subsets and combine them with 
a Logistic Regression.

\subsection{Feature Bagging}
Feature Bagging has been proposed by Lazarevic and Kumar \cite{lazarevicKumar2005Featurebagging} for multidimensional time series as a solution to anomaly detection problems in high-dimensional data. The authors argued that it is pointless to detect point anomalies based on the similarity (or distance) in high dimensions since the respective metrics lose their meaning when 
% as an anomaly detection method in multidimensional time series. The authors refer to algorithms that try to detect anomalies in high‐dimensional data, while at the same time, they consider it pointless to make such an effort because the importance of similarity (or distance) loses its meaning as 
the number of dimensions increases. According to the authors, 
% Due to the large number of dimensions the data are very far apart. This means that 
the larger the number of dimensions, the further the points are separated from each other, which results in multiple isolated points that are eventually (but falsely) considered anomalies. In addition, if we consider that most of the dimensions introduce noise, 
% Thus, the Feature Bagging method is proposed which takes a subset of the dimensions to avoid this phenomenon. On the other hand if we consider that most dimensions introduce noise, 
then it is more likely that anomalies will be found in a subset of the dimensions. So the Feature Bagging algorithm is proposed for creating subspaces, where it would be easier to identify anomalies.
% be used to be able to detect anomalies by looking at some of the dimensions and not all. 

The process for applying feature bagging and feeding the method ensemble is as follows:
\begin{enumerate}
    \item 
    Let $X=(X_t : t \in T)$ be a multivariate time series, where $T$ is the index set and $X_t \in \mathbb{R}^d, \forall t \in T$.
    % a data set, where $ X_i \in R^d $, with d being the number of dimensions of $X_i$.
    \item we select several basic (multivariate time series classification) algorithms (e.g. LSTM, autoencoders, etc.) and train a set $M$ of models. Individual models do not need to use the same architecture, more than one model can result from the same algorithm. For each model $ \text{Model}_m \in \{ \text{Model}_1, ... , \text{Model}_M \} $ we repeat the following steps:
    \item
        \begin{enumerate}
            \item We select a random number $N_m$, where $m$ denotes the $m-\text{Model}$, from a uniform distribution between $\lfloor \frac{d}{2} \rfloor$ and $(d-1)$. 
            % Here t denotes the t algorithm.
            \item Then we randomly select a subset $F_m$ of the original $X$ comprising $N_m$ features without repetition.
            % and create a subset F of features
            \item We train the anomaly detection model $\text{Model}_m$ on the $F_m$ subset.
            % the data keeping only the F features.
        \end{enumerate}
    \item for each instance, using the set of $\text{Model}_1, ... , \text{Model}_M$ models, we get an anomaly score $ASc_m$ from each model for this instance.
    % This method can stand by itself so at the end, we get for each point the anomaly score of the algorithm t $Anomaly\_Score_t$. So finally for each point we have T anomaly scores (one for each algorithm), $ASc_1, ASc_2, ..., ASc_t$ where $ASc_i$ is the Anomaly Score that the i-th model produced and we select
    \item The final anomaly score of an instance is generated by applying a collection function over the scores derived from each model, $ASc = agg(ASc_1, ASc_2, ..., ASc_M)$. We have selected Majority Voting as the collection function in our experiments.
\end{enumerate}

\subsection{Feature Bagging with Nested Rotations}
The technique has emerged from an algorithm that has been designed for classification problems, called Rotation Forest \cite{Rotation2006Forest}. Rotation Forest extends the popular Random Forest algorithm, which randomly selects subsets of the original feature space to train separate decision tree models, which are then combined in an ensemble. The main novelty of Rotation Forest is that it applies a PCA on the randomly selected subsets of features to get a "rotation matrix" (the principal components), which is then multiplied by the feature subsets to get the rotated features. The  algorithm is described below:
\begin{enumerate}
    \item First the data is normalized. Normalizing the data helps prevent very high values from dominating
and influencing the result.
    \item For each dimension the average value is calculated.
    \item The correlation matrix (covariance matrix) is then calculated for all dimensions. That is $\text{COV}(X,Y) = \frac{1}{m} \cdot \sum\limits_{i=1}^{m}(X-\mu_X)(Y-\mu_Y) $, $\forall X, Y$ where $X, Y$ are two dimensions (two features) and $X \neq Y$.
    %...X is and Y is and mu is......
    \item Then the eigenvectors and eigenvalues of the previous table are calculated using SVD. These vectors are also called Principal Components and the respective values represent the "importance" (or informativeness) of each component.
		\item We put the vectors in descending order according to the eigenvalues. That is, the vector with the largest eigenvalue is entered first, then the one with the next largest eigenvalue etc. Then we choose the $k$ first vectors.
		\item Finally, we take the matrix with the eigenvectors and multiply its inverse with the original data matrix to make the rotation, so:
		\[
		\begin{pmatrix}
			\text{new data}
		\end{pmatrix}
		=
		\begin{pmatrix}
			p_1 & ... & p_k
		\end{pmatrix} ^ T
		\begin{pmatrix}
			\text{data}
		\end{pmatrix}		
		\]
\end{enumerate}

Instead of keeping only the most "informative" eigenvectors we decide to keep all of them.
In doing so we rotate our data according to these axes. To create an ensemble using the rotation method the procedure is as follows:

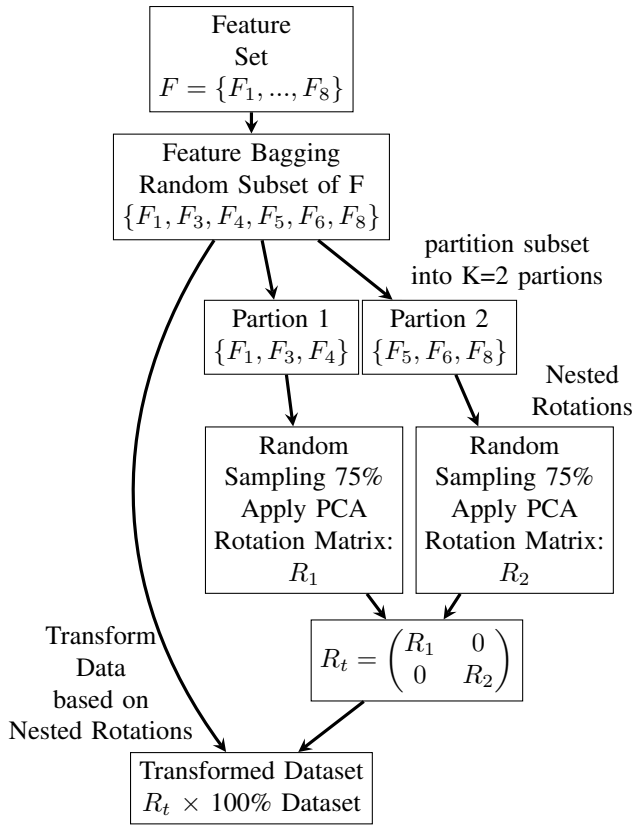
\begin{figure}
\centering
\begin{tikzpicture}[bend angle=35]
	
	\node (A1) [littlebox] at(0,0){Feature \\ Set \\ $F=\{F_1,...,F_8\}$};
	
	\node (B1)[littlebox] at($(A1) + (0,-1.7)$){Feature Bagging \\ Random Subset of F \\ $\{F_1,F_3,F_4,F_5,F_6,F_8\}$};
	\node (B2)[littlebox] at($(B1) + (0.4,-2)$){Partion 1 \\ $\{F_1,F_3,F_4\}$};
	\node (B3)[littlebox] at($(B2) + (2.1,0)$){Partion 2 \\ $\{F_5,F_6,F_8\}$};
	\node (B4)[littlebox] at($(B2) + (0.3,-2.3)$){Random \\ Sampling 75\% \\ Apply PCA \\ Rotation Matrix: \\ $R_1$};
	\node (B5)[littlebox] at($(B4) + (2.8,0)$){Random \\ Sampling 75\% \\ Apply PCA \\ Rotation Matrix: \\ $R_2$};
	\node (B6)[littlebox] at($(B4) + (1.5,-2)$){$R_t = \begin{pmatrix} R_1 & 0 \\ 0 & R_2 \end{pmatrix}$};
	\node (B7)[littlebox] at($(B1) + (0,-8)$){Transformed Dataset \\ $R_t$ $ \times $ 100\% Dataset};
	
	\node (C1) [nobox] at($(B3) + (0.9, 1)$){partition subset \\ into K=2 partions};
	\node (C2) [nobox] at($(B5) + (0.9,1.6)$){Nested \\ Rotations};
	\node (C3) [nobox] at($(B7) + (-2,1.4)$){Transform \\ Data \\ based on \\ Nested Rotations};

	\draw [arrow] (A1) -- (B1);
	\draw [arrow] (B1) -- (B2);
	\draw [arrow] (B1) -- (B3);
	\draw [arrow] (B2) -- (B4);
	\draw [arrow] (B3) -- (B5);
	\draw [arrow] (B4) -- (B6);
	\draw [arrow] (B5) -- (B6);
	\draw [arrow] (B6) -- (B7);
	
	\path[arrow, every node/.style={font=\sffamily\small}]
		(B1) edge[bend right] node [left] {} (B7);
		
\end{tikzpicture}
\caption{Nested Rotations} 
\label{fig:nested_rotations}
\end{figure}

\begin{enumerate}
    \item Let $X=(X_t : t \in T)$ be a multivariate time series, where $T$ is the index set and $X_t \in \mathbb{R}^d, \forall t \in T$.
    \item We select the $M$ basic models that will compose the ensemble model as before and for each of them we perform the following steps:
    \begin{enumerate}
        \item We apply the Feature Bagging technique by selecting a random number $N_m$ from a uniform distribution between $\lfloor \frac{d}{2} \rfloor$ and $(d-1)$ for each of the models. This step will give us a subset $F_m$. The $m$ index denotes the $m-\text{Model}$.
        \item for each subset $F_m$ we apply a transformation based on nested rotations by further partitioning it into $K$ subsets such that $K_{m_1} \bigcup K_{m_2} \bigcup ... \bigcup K_{m_K} = F_m$. We then apply PCA on these subsets as shown in Figure \ref{fig:nested_rotations}. So each subset $F_m$ is further partitioned into $K$ partitions:
        \begin{itemize}
            \item We subsampling the subset $F_m$ in order to increase diversity of the partitions. This is done because if two models $m1, m2$ happens to have the same subset $F_{m1}=F_{m2}$ and happens to have the same partitioning $\{f_1, f_2, f_3\} $ and $ \{f_4, f_5, f_6\} $ then by subsampling the $F_{m1}=F_{m2}$ resulting into different partitions and thus $K_{m2_1} \neq K_{m1_1}$ and $K_{m2_2} \neq K_{m1_2}$. So by applying PCA (in the next steps) it is not resulting in the same rotation matrices.
            \item We apply PCA on each partition.
            \item from each partition we get a rotation matrix $R^m_k$, where $k$ denotes the $k$-partition with dimensions $(\text{dimension of } F_m)/K \times (\text{dimension of } F_m)/K$ (for simplicity we suppose that the dimension of $F_m$ is divided with $K$ but it is not necessary).
        \end{itemize}
        \item From the previous step (which we called Nested Rotations) we have a transformation for each subset $F_m$: 
        {
        \small
		\[
		R_m = 
		\begin{pmatrix}
			R^m_1 & 0 & \cdots & 0 \\
			0 & R^m_2 & \cdots & 0 \\
			\vdots & \vdots & \ddots & \vdots \\
			0 & 0 & \cdots & R^m_K
		\end{pmatrix}		
		\]
        }
        where each $R^m_k$ denotes a submatrix corresponding to a partition and 0 denotes the zero-submatrix.
        \item Finally, we get the new (nested rotated) data by applying $R_m$ on the subset $F_m$. So each the models $\text{Model}_1, ..., \text{Model}_M$ is trained on these "nested rotated" data.
    \end{enumerate}
    \item Finally for each instance, using the set of $\text{Model}_1, ... , \text{Model}_M$ models, we get an anomaly score $ASc_m$ from each model for this instance.
    \item The final anomaly score of an instance is generated by applying a collection function over the scores derived from each model, $ASc = agg(ASc_1, ASc_2, ..., ASc_M)$. We have selected Majority Voting as the collection function in our experiments.
\end{enumerate}

The number of individual models M, the number of partitions K that we split each subset $F_m$ the portion of data that we use when we subsampling the each $F_m$ to in order to create the partitions, all these are parameters of this algorithm.

\subsection{Stacking Feature Bagging with Nested Rotations models}
In this section, we make use of the previous technique to boost the performance of our prediction in a semi-supervised setup using a Logistic Regressor to combine the individual models as a collection function instead of the Majority Voting we used in the unsupervised setup.

Let $X=(X_t : t \in T)$ be a multivariate time series, where $T$ is the index set and $X_t \in \mathbb{R}^d, \forall t \in T$.
We select the $M$ basic models that will compose the ensemble model as before and for each of them we perform the following steps: The procedure for implementing this ensemble method is as follows:
\begin{enumerate}
    \item First, we divide the training set into two training subsets, one (subset A) for training the individual models and one (subset B) for training the Logistic Regressor. This split guarantees that there will be no leak of information between the individual models and the Logistic Regressor.
    \item In the next step we choose the number of individual models M we need to create using the Feature Bagging technique combined with Nested Rotations as in the previous section.
    \item With these individual models (trained on data set A) we make predictions on the data set $B$ that the models have not been trained on. 
    \item As a final step we train a Logistic Regression to improve the prediction performance. More precisely a prediction is made on every element of data set $B$, and thus every element gets T anomaly scores. So a new data set created $B' = {b_1, b_2, b_3} $ where $b_i = (ASc^i_1, ASc^i_2,...,ASc^i_T)$ where $ASc^i_t$ is the anomaly score that model $\text{Model}_m$ gave to this element $b_i$. So at the final semi-supervised step, we train a Logistic Regression model in data set $B'$ given the labels, to learn how to combine the individual predictions.
\end{enumerate}

The number of individual models M, the number of partitions K that we split each subset $F_m$ the portion of data that we use when we subsampling the each $F_m$ to in order to create the partitions, all these are parameters of this algorithm which derived from the Feature Bagging with Nested Rotations algorithm.

\section{Experimental Evaluation}

\subsection{Experimental setup}
% , dataset, and metrics}
The proposed models of anomaly detection in multivariate time series with ensemble techniques are implemented and evaluated in Python programming language using the frameworks NumPy, pandas, Scikit-learn, TensorFlow 2, and the Keras API. 
Specifically, we used five basic architectures which are Autoencoder, LSTM, LSTM Autoencoder, LSTM Variational Autoencoder, and Convolutional Autoencoder.
The experiments took place in the Google Colab environment except for the model that was proposed in \cite{elhalwagy2022hybridization} called LSTMcaps.

The datasets we used are taken from Skoltech Anomaly Benchmark (SKAB) \cite{skabdataset}, which is a collection of time series produced by the operation of a water pump device (motor) monitored by various sensors. The set of sensors generates eight values at every moment thus resulting in multivariate time series with eight features, and each monitoring file contains normal operation and anomaly points. SKAB therefore can be used to evaluate techniques and models in the context of Anomaly Detection research in multivariate time series. SKAB contains 34 multivariate time series of 37401 moments in total or 1100.02 moments per time series on average. Their size is 3.41 MB in total or 102.76 KB per time series on average. They do not contain missing values and they do not contain duplicates. Each point of the time series is a vector of 10 values which are, the timestamp (representing the exact time when the following values were recorded), the anomaly label, and the 8 values that sensors recorded (Accelerometer1RMS, Accelerometer2RMS, Current, Pressure, Temperature, Thermocouple, Voltage, and Volume Flow RateRMS). Each time series is a recorded experiment that starts in a normal state and after a while by switching valves, anomalies are injected.
%\todo[inline]{Give 1 or 2 sentences describing some characteristics of the dataset. For instance, the size in terms of MB, the number of time series, how you made the split training/testing, if we had misses/duplicate values. Describe the eight features and the time series i.e. water flow on a water circuit}

% \textbf{[Describe the evaluation metrics you use, 1 paragraph]}
The evaluation metrics that we report are the F1-Score and Area Under Curve (AUC) of the receiver operating characteristic curve (ROC curve), which is more appropriate for a task that is by definition highly unbalanced between the majority  (normal) and the minority (anomaly) class. 

The AUC is defined as follows: 
Let $\text{TPR}=\frac{TP}{P}$ and $\text{FPR}=\frac{FN}{N}$ be the true positive rate and the false positive rate respectively, where TP stands for the true positive cases, P for all the positive cases, FN for the false negatives and N for all the negative cases. The anomalous points are considered positives and the non-anomalous points are considered negatives. Then the AUC metric is the area under the curve defined by ROC curve as plotted from TPR and FPR pairs for various $\delta \in \mathbb{R}$ where $\delta$ is the threshold of our anomaly detection method. F1-Score defined as $ \text{F}_1 = \frac{2 \cdot \text{Precision} \cdot \text{Recall}}{\text{Precision}+\text{Recall}} $ where $ \text{Precision} = \frac{\text{TP}}{\text{TP}+\text{FP}}, \quad \text{Precision} = \frac{\text{TP}}{\text{TP}+\text{FN}} $. For both metrics, higher values denote better performance. These two metrics are widely used to evaluate anomaly detection methods (e.g. in \cite{skabdataset, elhalwagy2022hybridization} and in \cite{munir2018deepant, goldstein2016comparative}). 

% [If we need more text]:
% On the other hand the traditional accuracy which definced as Accuracy = (TN + TP) / (TP + FP + TN + FN) is not suitable in this problem. A simple example to think of it is that if we have 100 points with 99 of them be normal and only 1 be anomalous, then if our method fails to find the anomaly the accuracy would have show us that the performance is 99\% which is confusing because we missed the only one anomaly our data set contains.

\subsection{Results}
In Table \ref{tab:modelcomparison} we can see the comparative results of the proposed ensemble methods and other baseline approaches. Every section consists of the architecture the ensemble is based on. We further divide each section and show the F1 Score and AUC regarding 3 types of models of the corresponding architecture, the plain version of the model, the ensemble using Feature Bagging, and the ensemble using Feature Bagging with Nested Rotations. We also included LSTMcaps using the F1 Score that the authors provided in their work. Furthermore, we extend the table and added a section on top of the table for our semi-supervised approach.
For each architecture and for each time series we created and train each model as follows:
\begin{enumerate}
    \item Split/slice the time series into two parts.
    \item Train the corresponding model with the first part of the time series. For the "mixed" section we used 3 parts, the first two for training and the third part for testing as described in the subsection "Stacking Feature Bagging with Nested Rotations models".
    \begin{enumerate}
        \item for the Feature Bagging model we created 17 models of the corresponding architecture.
        \item for the Feature Bagging with Nested Rotations model we created 17 models of the corresponding architecture. For the parameters of all models, we set the proportion of data to be 75\%, and the K partitions on which the PCA algorithm is applied was set to 2.
        \item for the architecture "mixed" we used 60 Feature Bagging with Nested Rotations (12 for each architecture) using the first part of the time series and combined them with a Logistic Regressor using the second part of the time series. For the parameters of all models, we set the proportion of data to be 75\%, the K subset on which the PCA algorithm is applied was set to 2.
    \end{enumerate}
    \item Test the corresponding model using the second part of the time series:
    \begin{enumerate}
        \item for the Feature Bagging model of the corresponding architecture every instance/moment of the time series has 17 labels of 0/1 and we use Majority Voting as an aggregation function to conclude to a single 0/1 label.
        \item for the Feature Bagging with Nested Rotations model of the corresponding architecture using the same logic as above.
        \item for the architecture "mixed" we used the third part of the time series for testing. 
    \end{enumerate}
\end{enumerate}
At the end of this process, we use the macro-average to conclude the F1 Score and AUC for every architecture and its corresponding models. This evaluation approach has been followed by other researchers in similar tasks (e.g. \cite{skabdataset} and \cite{elhalwagy2022hybridization}) to evaluate their models.

In Table \ref{tab:modelcomparison} that follows, we see that Feature Bagging can slightly improve the performance of most base methods. More specifically, the use of Feature Bagging improved the best performance of the plain versions of autoencoder, LSTM autoencoder and LSTM VAE, whereas couldn't beat the best performance of the plain versions of a convolutional autoencoder and the LSTM model without Feature Bagging. This is an indication that Feature Bagging captures anomalies even if they are hidden in a subset of features. To further increase the performance then we combine Feature Bagging and Nested Rotations and again we combine the results with majority voting. We can clearly see that these two techniques in combination provide us with around 2-4\% better performance than the plain version of models. The only model that beats the performance of Feature Bagging with Rotation is the LSTM.

As we can see the best model in the unsupervised setup is when we used the Convolutional Autoencoder architecture applying Feature Bagging with Nested Rotations techniques resulting in an ensemble using Majority Voting. This model has a 0.7873 F1 Score and 0.8315 AUC which outperforms the others. Finally as expected in the semi-supervised environment the performance is boosted, and the improvement is at least 10\%.

\begin{table}[!htb]
    \centering
    \caption{Model Comparison}
    \begin{tabular}{|>{\centering\arraybackslash}m{2cm} |>{\centering\arraybackslash}m{3cm}|>{\centering\arraybackslash}m{1cm}|>{\centering\arraybackslash}m{1cm}|}
        
        \hline
        Architecture (mode) & Model Title & F1 Score & AUC \\
        \hline
        Mixed (semi-supervised) & Stacking FBR models with Logistic Regression & \textbf{0.85} & \textbf{0.88} \\
        \hline
        
        \specialrule{0em}{.25em}{.25em}

        \hline
        \rule{0pt}{4ex}
        \multirow{3}{=}{\centering Convolutional Autoencoder (unsupervised)}& \textbf{Feature Bagging with Nested Rotations} & \textbf{0.7873}  & \textbf{0.8315} \\
        
        \cline{2-4}
        \rule{0pt}{4ex}
        & Feature Bagging & 0.7451 & 0.80 \\
        \cline{2-4}
        \rule{0pt}{4ex}
        & Plain & 0.7622 & 0.8117 \\
        \hhline{|=|=|=|=|}
        \rule{0pt}{4ex}
        \multirow{3}{=}{\centering LSTM Autoencoder (unsupervised)}& \textbf{Feature Bagging with Nested Rotations} & \textbf{0.7641} & \textbf{0.8190} \\ 
        \cline{2-4}
        \rule{0pt}{4ex}
        & Feature Bagging & 0.7465 & 0.8050 \\
        \cline{2-4}
        \rule{0pt}{4ex}
        & Plain & 0.7410 & 0.8021 \\
        \hhline{|=|=|=|=|}
        \rule{0pt}{4ex}
        LSTMCaps (unsupervised) & LSTMCaps & 0.74  & - \\
        \hhline{|=|=|=|=|}
        \rule{0pt}{4ex}
        \multirow{3}{=}{\centering LSTM (unsupervised)}& Feature Bagging with Nested Rotations & 0.7074 & 0.7749 \\ 
        \cline{2-4}
        \rule{0pt}{4ex}
        & Feature Bagging & 0.6723 & 0.7500 \\
        \cline{2-4}
        \rule{0pt}{4ex}
        & Plain & \textbf{0.7225} & \textbf{0.7867} \\
        \hhline{|=|=|=|=|}
        \rule{0pt}{4ex}
        \multirow{3}{=}{\centering LSTM Variational Autoencoder (unsupervised)}& \textbf{Feature Bagging with Nested Rotations} & \textbf{0.7014} & \textbf{0.7704}   \\
        \cline{2-4}
        \rule{0pt}{4ex}
        & Feature Bagging & 0.6978 & 0.7680 \\
        \cline{2-4}
        \rule{0pt}{4ex}
        & Plain & 0.6815 & 0.7570 \\
        \hhline{|=|=|=|=|}
        \rule{0pt}{4ex}
        \multirow{3}{=}{\centering Autoencoder (unsupervised)}& \textbf{Feature Bagging with Nested Rotations} & \textbf{0.6259} & \textbf{0.7231} \\
        \cline{2-4}
        \rule{0pt}{4ex}
        & Feature Bagging & 0.5999  & 0.7089 \\
        \cline{2-4}
        \rule{0pt}{4ex}
        & Plain & 0.5935 & 0.7050 \\
        \hline
    \end{tabular}
    \label{tab:modelcomparison}
\end{table}

\subsection{Discussion}
From the results of the tables \ref{tab:modelcomparison}, we see that Feature Bagging can improve the performance of some models but not all of them. However, when it is combined with nested rotations, the performance is usually better.
% This leaves room for us to add the other technique.
% Guesses:
% 1. in high dimensinality it may performs much better
% 2. it is more robust in the noise. Our data do not contain too much noise. (i think there is not noise at all)
More specifically, when we combine it with the transformation based on nested rotations computed by PCA into the Feature Bagging with Nested Rotations technique we have an overall improvement of 2\%. The fact that every Feature Bagging with Nested Rotations performs better against the majority of the corresponding architectures leads us to the conclusion that this ensemble technique can take anomaly detection 
% based on these architectures 
a step further. Also, we can conclude that Feature Bagging and Nested Rotations act complementary: Feature Bagging reveals the anomalies and the features that are most affected, whereas Nested Rotations inject diversity and boost the performance of the ensemble. Integrating multiple PCAs with feature bagging leads to increased effectiveness in most cases, as can be seen from the results in Table \ref{tab:modelcomparison}. One possible explanation for this is that multiple PCAs after random subspace selection further increases the diversity of the individual classifiers in the ensemble. A deeper investigation of the effect of PCA on the model ensemble is left as future work.

On the other hand, the semi-supervised ensemble model which combines Feature Bagging with Nested Rotations on the individual models, using a Logistic Regressor, gives us the best results on the anomaly detection problem and outperforms all other methods. However, its main limitation is that it needs a few training samples in order to learn how to combine individual models. It also needs much more time for training and inference since it produces multiple models. This is due to the fact that apart from the 60 learners it employs, the PCA algorithm is very time-consuming in its computations. Since PCA is applied many more times than the number of models in the ensemble the computational cost can significantly increase. However, the process can be easily parallelized using a new thread for each model and a multi-core system to do the training or inference.
% So it needs some time to train individual models and on top of that it PCA adds some extra time in it. However the design has its own advantage.  The benefit is that each learner can be trained in parallel. Learners do not depend on each other so they can be trained in parallel in a multi core system and thus the model needs much less time to be trained.

About the ensemble models (either the unsupervised or the one in semi-supervised) the final number of learners as well as the other parameters of algorithms has been chosen after performing a grid-like search that balances between the available architectures and the number of models to train for each architecture and the other parameters while respecting our limitations in time and computational resources. 
%Sixty models in total have been trained and combined in the ensemble, in an attempt to maximize prediction performance while respecting our limitations in time and computational resources. 
% Actually we refer to the term grid-like because we couldn't test much larger values due to the limitation of resources and time as we discussed above. Finally 

\section{Conclusions and Next steps}

In this work, we implemented two ensemble techniques applying in unsupervised mode and semi-supervised mode for anomaly detection in multivariate time series. This problem has been approached by various algorithms both from machine techniques and deep learning models. We used 5 basic deep machine learning models and built on them to implement ensemble techniques to achieve even better results. The ensemble models built initially were using two techniques called Feature Bagging and a transformation based on Nested Rotations computed by PCA. The result showed us that these techniques could improve our basic models in combination with unsupervised learning. Feature Bagging alone had almost the same performance as basic models in some cases. When we combined it with Nested Rotations using PCA algorithm, in most cases, performed better than the basic models. Finally, when we combined all of the above into a general model called Stacking/Fusion Feature Bagging with Rotation using a Logistic Regressor in semi-supervised mode and we got the best performance from all the previous techniques as expected. Ensemble techniques are known to greatly improve performance and solve many problems and our results showed that they can also help a lot in detecting anomalies.

% \textbf{As feature work we plan }
Although these models are performing quite well in the anomaly detection problem in an unsupervised or semi-supervised environment they heavily depend on the aggregation function applied to the multiple scores we get at the final stage. Hence it is a valid question whether another function or even a non-linear model can boost the performance further. In \cite{shahzad2013comparative} they did research on variants of majority voting and presented the two veto-based voting schemes that can combine multiple classifiers together based on their reliability of them. Another important point for future work is to test our model in higher dimensions. The two techniques we used Feature Bagging and Rotation are the core of our models. As we discussed earlier Feature Bagging is robust in high dimensions while Rotation injects diversity and boosts performance. Thus, we believe that as the dimensionality increases, the model should not be affected, but more tests with time series in higher dimensions need to be done to confirm this. Furthermore, as we already mentioned PCA seems to help a lot but a deeper investigation of the effect of PCA on the model ensemble is left as future work.
%Additionally, this can be extended in a way that the model does not need labels and move the model under unsupervised training. 

% Anomaly score normalization
% \section*{Acknowledgment}

%  This work is part of the X project that has received funding from the European Union’s Horizon 2020 research and innovation programme under grant agreement No X".

\section*{Acknowledgment}
The work leading to these results has received funding from the European Union’s Horizon 2020 research and innovation programme under Grant Agreement No. 965231, project REBECCA (REsearch on BrEast Cancer induced chronic conditions supported by Causal Analysis of multi-source data).

\bibliographystyle{IEEEtran}
\bibliography{references}

\end{document}